\documentclass{article}

 \usepackage[preprint]{neurips_2025}

\usepackage[utf8]{inputenc} 
\usepackage[T1]{fontenc}    
\usepackage{hyperref}       
\usepackage{url}            
\usepackage{booktabs}       
\usepackage{amsfonts}       
\usepackage{nicefrac}       
\usepackage{microtype}      
\usepackage{xcolor}         

\newcommand{\eg}{\textit{e.g.}}
\newcommand{\ie}{\textit{i.e.}}

\usepackage{pifont}
\newcommand{\cmark}{\ding{52}}%
\newcommand{\xmark}{\ding{56}}%
\usepackage{graphicx}
\newcommand{\hcmark}{\ding{52}\rotatebox[origin=c]{-9.2}{\kern-0.7em\ding{55}}}
\usepackage{subcaption}
\usepackage[most]{tcolorbox}
\usepackage{listings}
\usepackage{multirow}

\usepackage{forest}
\forestset{
  default preamble={
    for tree={
      grow'=0,
      draw,
      rounded corners,
      align=left,
      parent anchor=east,
      child anchor=west,
      edge={->},
      l sep+=10pt,
      s sep+=6pt,
      font=\footnotesize
    }
  }
}

\title{Surg$\Sigma$: A Spectrum of Large-Scale Multimodal Data and Foundation Models for Surgical Intelligence}

\author{
  Zhitao Zeng$^{1}$\thanks{These authors contributed equally to this work. \quad $^\dagger$Corresponding authors.} \quad  Mengya Xu$^{2*}$ \hspace{0.4em} Jian Jiang$^{3*}$ \hspace{0.4em} Pengfei Guo$^{4*}$ \hspace{0.4em} Yunqiu Xu$^{1}$ \hspace{0.4em} Zhu Zhuo$^{1}$\\
    \textbf{Chang Han Low$^{1}$ \hspace{0.4em}  Yufan He$^{4}$ \hspace{0.4em}  Dong Yang$^{4}$ \hspace{0.4em} Chenxi Lin$^{3}$ \hspace{0.4em} Yiming Gu$^{3}$ \quad Jiaxin Guo$^{2}$}\\
    \textbf{Yutong Ban$^{3\dagger}$ \qquad  Daguang Xu$^{4\dagger}$ \qquad  Qi Dou$^{2\dagger}$ \qquad  Yueming Jin$^{1\dagger}$}\\
  $^{1}$NUS \qquad $^{2}$CUHK \qquad $^{3}$SJTU \qquad $^{4}$NVIDIA\\
  Project page: \url{https://SurgSigma.github.io} \\
}

\begin{document}

\maketitle

\begin{abstract}
Surgical intelligence has the potential to improve the safety and consistency of surgical care, yet most existing surgical AI frameworks remain task-specific and struggle to generalize across procedures and institutions. 
Although multimodal foundation models, particularly multimodal large language models, have demonstrated strong cross-task capabilities across various medical domains, their advancement in surgery remains constrained by the lack of large-scale, systematically curated multimodal data.
To address this challenge, we introduce Surg$\Sigma$, a spectrum of large-scale multimodal data and foundation models for surgical intelligence.
At the core of this framework lies Surg$\Sigma$-DB, a large-scale multimodal data foundation designed to support diverse surgical tasks.
Surg$\Sigma$-DB consolidates heterogeneous surgical data sources (including open-source datasets, curated in-house clinical collections and web-source data) into a unified schema, aiming to improve label consistency and data standardization across heterogeneous datasets.
Surg$\Sigma$-DB spans 6 clinical specialties and diverse surgical types, providing rich image- and video-level annotations across 18 practical surgical tasks covering understanding, reasoning, planning, and generation, at an unprecedented scale (over 5.98M conversations).
Beyond conventional multimodal conversations, Surg$\Sigma$-DB incorporates hierarchical reasoning annotations, providing richer semantic cues to support deeper contextual understanding in complex surgical scenarios.
We further provide empirical evidence through recently developed surgical foundation models built upon Surg$\Sigma$-DB, illustrating the practical benefits of large-scale multimodal annotations, unified semantic design, and structured reasoning annotations for improving cross-task generalization and interpretability.
 \end{abstract}

\section{Introduction}
According to estimates from the Lancet Commission, more than 300 million surgical procedures are performed worldwide each year~\cite{meara2015global}, underscoring the urgent demand for safer and more accessible surgical care. 
Despite advances in minimally invasive~\cite{ferrari2024death,mussa2025single} and robotic techniques~\cite{dupont2021decade,jiang2025current}, surgery remains inherently complex, requiring continuous interpretation of dynamic anatomy and high-stakes decision-making under uncertainty.
Surgical AI is therefore emerging as a transformative paradigm, acting as an intelligent collaborator that enhances perception, understanding, and reasoning. 
By leveraging multimodal intraoperative signals (\eg, visual streams, textual instructions, robotic kinematics, and preoperative imaging), AI systems promise to improve safety, reduce variability, and broaden access to high-quality surgical expertise. 
However, most prior surgical AI systems remain narrowly designed for isolated tasks, including phase recognition~\cite{twinanda2016endonet,lavanchy2024challenges}, tool or tissue segmentation~\cite{allan20192017,allan20202018}, and action classification~\cite{psychogyios2023sar,ayobi2025pixel}, often within a tailored task or a single surgical type. 
This task-specific paradigm limits knowledge transfer and leads to brittle generalization, where models degrade under distribution shifts caused by differences in imaging systems, anatomy, or surgical styles.

Foundation models, particularly multimodal large language models~\cite{team2023gemini,singh2025openai,bai2025qwen25vltechnicalreport,bai2025qwen3}, have recently emerged as a promising paradigm for unified visual perception, language understanding and multimodal reasoning, enabling models to jointly interpret visual content and reason with natural language. 
In principle, such models offer a unified framework capable of supporting a broad spectrum of surgical tasks, ranging from describing anatomical structures and instrument states to answering intraoperative queries, summarizing procedural context, and generating interpretable decision-support rationales. 
While foundation models have achieved remarkable success across domains such as radiology~\cite{wu2025towards,sun2025foundation,bluethgen2025vision}, pathology~\cite{chen2024towards,dingmultimodal,vorontsov2024foundation}, and molecular biology~\cite{jumper2021highly,abramson2024accurate}, their application to the surgical domain remains comparatively underexplored.
Surgery poses fundamentally distinct challenges for training multimodal foundation models. 
Intraoperative scenes are not only visually complex (\eg, severe occlusion, tissue deformation, and rapid camera motion), but also exhibit strong spatiotemporal structure and causal interdependence, where subtle instrument–tissue interactions can induce irreversible anatomical changes.
Clinically relevant cues are often fine-grained, transient, and context-dependent, demanding long-horizon temporal reasoning and precise spatial grounding beyond static image understanding. 
Furthermore, variability across institutions, surgeons, devices, and patient anatomies introduces substantial distribution shifts that hinder generalization.

We observe that a fundamental obstacle to advancing surgical multimodal foundation models lies in the lack of large-scale, high-quality, and systematically curated multimodal data. 
Conventional surgical datasets~\cite{7519080,ayobi2025pixel,wang2022autolaparo} are predominantly vision-centric and designed under a closed-set paradigm, providing only predefined categorical annotations (\eg, surgical phase or instrument tags) while lacking practical natural language instruction–visual pairs that better reflect real-world clinical usage and flexible interaction. 
In addition, these datasets are typically limited in scale and surgical-type diversity, as they are often confined to a small number of procedures or institutions. 
Consequently, \textbf{they remain fragmented across tasks and modalities, with inconsistent annotation standards and heterogeneous label spaces that hinder cross-dataset integration and large-scale training}. 
Although some recent works~\cite{cheng2025benchmarking,sureon2026,qin2026surgo} have introduced datasets for surgical foundation models, they still exhibit notable limitations in scale, diversity, and task coverage, as summarized in Table~\ref{tab: dataset compare}. 
On the other hand, annotation quality and granularity remain insufficient: the absence of a unified label space can mislead training and weaken generalization, while existing datasets largely lack high-quality multi-step reasoning traces.

\begin{figure}[t]
  \centering
  \includegraphics[width=\linewidth]{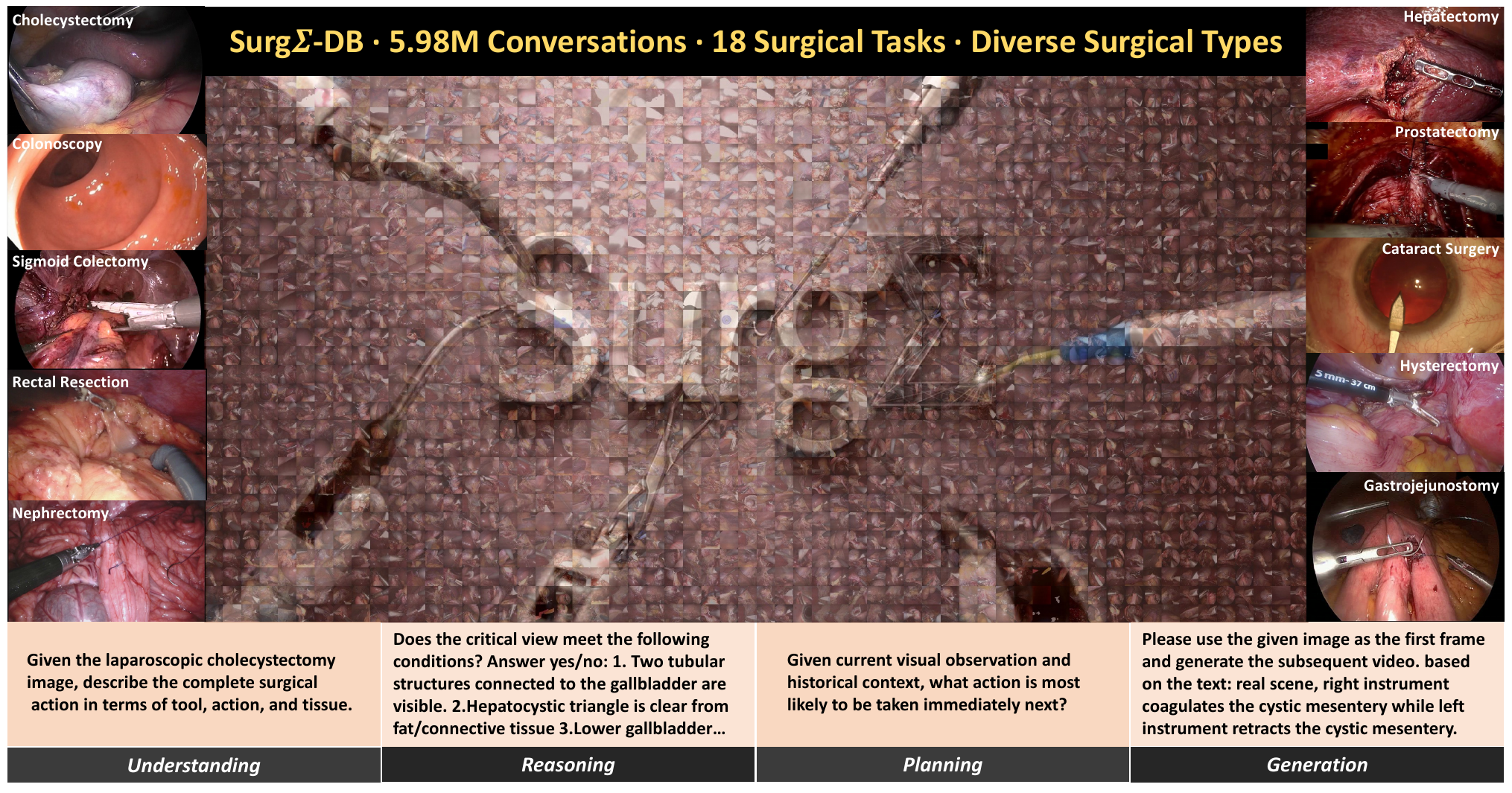}
  \vspace{-1.5em}
  \caption{Surg$\Sigma$-DB is a large-scale multimodal data foundation for surgical intelligence. }
  \vspace{-1em}
  \label{fig: teaser}
\end{figure}

To facilitate research, we present Surg$\Sigma$, a spectrum of large-scale multimodal  data and foundation models for surgical intelligence. 
At its core, Surg$\Sigma$-DB serves as a unified multimodal data foundation designed to enable large-scale training of surgical foundation models. 
Rather than releasing isolated task-specific datasets, our Surg$\Sigma$-DB systematically consolidates diverse surgical data sources into a unified and well-structured foundation.
We curate data spanning multiple surgical specialties and procedures, covering diverse clinical departments and operative types to ensure broad anatomical and procedural variability. 
\textbf{Surg$\Sigma$-DB combines open-source resources with web-collected surgical videos, and employs a semi-automated annotation pipeline integrating expert human labeling and controlled synthesis} to ensure both real-world representativeness and scalable clinical fidelity.
Within the same surgical scenes, \textbf{Surg$\Sigma$-DB provides rich annotations with hierarchical reasoning traces}, enabling multi-grained spatial understanding as well as temporal modeling. 
To the best of our knowledge, \textbf{Surg$\Sigma$-DB provides one of the most comprehensive task coverages in surgical intelligence}, spanning understanding, reasoning, planning, and generative capabilities through richly structured and multi-level annotations, and \textbf{is constructed at an unprecedented scale} (\ie, $\sim$5.98M) across diverse surgical types from 6 clinical specialties.
Importantly, all data are organized under a unified format, facilitating interoperable training, cross-task integration, and future extensibility, while promoting more consistent label spaces across heterogeneous datasets.

Building upon Surg$\Sigma$-DB, a family of surgical foundation models (\ie, BSA~\cite{xu2026bsa}, SurgVLM~\cite{zeng2025surgvlm}, Surg-R1~\cite{jiang2026surgr1}, and Cosmos-H-Surgical~\cite{he2025surgworld}) are developed, which empirically demonstrate the effectiveness of the proposed unified data spectrum.
BSA~\cite{xu2026bsa} demonstrates that fundamental surgical actions exhibit consistent, recognizable patterns across anatomically diverse procedures, enabling cross-specialty generalization without domain-specific adaptation and supporting clinically meaningful downstream applications including skill assessment and procedural planning.
SurgVLM~\cite{zeng2025surgvlm} demonstrates that large-scale multimodal instruction-tuning data can enhance cross-task generalization, enabling a single model to effectively handle diverse surgical understanding tasks through shared vision–language training.
Surg-R1~\cite{jiang2026surgr1} further illustrates the critical role of structured reasoning annotation, where multi-step inference traces significantly strengthen grounded surgical understanding. 
Cosmos-H-Surgical~\cite{he2025surgworld} demonstrates that surgical world models can transform large-scale unlabeled surgical video into actionable training data for robot policy learning by synthesizing realistic surgical scenes and recovering pseudo-kinematics through inverse dynamics inference, thereby enabling scalable vision–language–action training with limited real demonstrations and significantly improving policy performance and sample efficiency.
Together, these models provide complementary evidence that scale, semantic unification, and chain-of-thought reasoning annotations are key ingredients for advancing surgical foundation models within a coherent data-centric framework.

We hope that our data foundation and preliminary findings will inspire the research community to further explore and unlock the untapped potential of multimodal foundation models in surgical intelligence, ultimately advancing clinically reliable and generalizable surgical AI systems.
In future work, we will continually expand Surg$\Sigma$-DB in data scale and diversity, and progressively enrich each surgical scene with comprehensive and holistic annotations toward full task coverage within unified surgical contexts.
In summary, our contributions can be summarized as follows:
\begin{itemize} 
\item \textbf{A large-scale multimodal surgical data foundation.}  
We introduce Surg$\Sigma$-DB, a large-scale multimodal dataset spanning multiple surgical specialties and procedures. The dataset integrates image- and video-level annotations across understanding, reasoning, planning, and generation tasks, providing the most comprehensive task coverages in surgical intelligence.

\item \textbf{Comprehensive and unified multi-granular annotations.}  
We consolidate heterogeneous data sources into a unified data schema with a consistent label space. 
A semi-automated annotation pipeline combining human labeling and controlled synthesis with hierarchical reasoning annotations enables semantic coherence and scalable foundation model training. 

\item \textbf{Empirical validation through foundation models.}  
A family of surgical foundation models is developed upon Surg$\Sigma$-DB, providing empirical validation of key data design principles and demonstrating the impact of large-scale multimodal data foundation.

\end{itemize}

\begin{table}[t]
\scriptsize
\caption{Comparison with existing multimodal surgical datasets and benchmarks.}
\label{tab: dataset compare}
\centering
\setlength{\tabcolsep}{1.5pt}
\begin{tabular}{lcccccccccccc}
\toprule
\multirow{2}{*}{\textbf{Datasets}} 
& \multicolumn{2}{c}{\textbf{Visual Modality}} 
& \multicolumn{3}{c}{\textbf{Conversation Type}} 
& 
& \multicolumn{3}{c}{\textbf{Data Source}} 
& \multirow{2}{*}{\textbf{Reasoning}} 
& \multirow{2}{*}{\textbf{\#Sample}} 
& \multirow{2}{*}{\textbf{\#Task}} \\
\cmidrule(r){2-3} \cmidrule(r){4-6} \cmidrule(r){8-10}
 & \textbf{Video} & \textbf{Image} 
 & \textbf{VQA} & \textbf{Caption} & \textbf{Generation} 
 & 
 & \textbf{In-House} & \textbf{Open-Source} & \textbf{Internet} 
 &  &  &  \\
\midrule
Cholec80-VQA~\cite{seenivasan2022surgical} & \xmark & \cmark & \cmark & \xmark & \xmark & & \xmark & \cmark & \xmark & \xmark & 43K & 3 \\
EndoVis2018-VQA~\cite{seenivasan2022surgical} & \xmark & \cmark & \cmark & \xmark & \xmark & & \xmark & \cmark & \xmark & \xmark & 11.78K & 3 \\
PSI-AVA-VQA~\cite{seenivasan2023surgicalgpt} & \xmark & \cmark & \cmark & \xmark & \xmark & & \xmark & \cmark & \xmark & \xmark & 10.29K & 3\\
PitVQA~\cite{PitVQA} & \xmark & \cmark & \cmark & \xmark & \xmark & & \xmark & \cmark & \xmark & \xmark & 884.24K & 5\\
LRSP-VQA~\cite{chen2024llm} & \xmark & \cmark & \cmark & \xmark & \xmark & & \xmark & \xmark & \cmark & \xmark & 1.13K & 4 \\
CoPESD~\cite{wang2025copesd} & \xmark & \cmark & \cmark & \xmark & \xmark & & \xmark & \cmark & \xmark & \xmark & 121.09K & 3\\
EndoVQA-Instruct~\cite{liu2025endobench} & \xmark & \cmark & \cmark & \xmark & \xmark & & \cmark & \cmark & \xmark & \xmark & 446.54K & 12 \\
Surg-396K~\cite{wang2025endochat} & \xmark & \cmark & \cmark & \cmark & \xmark & & \xmark & \cmark & \xmark & \xmark & 396K & 7 \\
SurgPub-Video~\cite{li2025surgpub} & \cmark & \xmark & \cmark & \xmark & \xmark & & \xmark & \cmark & \cmark & \xmark & 48.52K & 3 \\
SurgMLLMBench~\cite{choi2025surgmllmbench} & \xmark & \cmark & \cmark & \xmark & \xmark & & \xmark & \cmark & \xmark & \xmark & 893.15K & 5 \\
SurgLaVi~\cite{perez2025surglavi} & \xmark & \cmark & \cmark & \xmark & \xmark & & \cmark & \cmark & \xmark & \xmark & 239.8K & 4 \\
SurgVeo~\cite{chen2025far} & \cmark & \cmark & \xmark & \xmark & \cmark & & \xmark & \cmark & \xmark & \xmark & 50 & 1 \\
SVU-31K~\cite{wang2025surgvidlm} & \cmark & \xmark & \cmark & \xmark & \xmark & & \xmark & \cmark & \cmark & \xmark & 31K & 4 \\
SurgCoTBench~\cite{low2026surgraw} & \xmark & \cmark & \cmark & \xmark & \xmark & & \xmark & \xmark & \cmark & \cmark & 14.25K & 5 \\
SUREON~\cite{sureon2026} & \cmark & \xmark & \cmark & \xmark & \xmark & & \xmark & \cmark & \xmark & \cmark & 206.8K & 12 \\
\midrule
Surg$\Sigma$-DB (ours) & \cmark & \cmark & \cmark & \cmark & \cmark & & \cmark & \cmark & \cmark & \cmark & 5.98M & 18 \\
\bottomrule
\end{tabular}
\vspace{-1em}
\end{table}

\section{Related Work}
\subsection{Surgical Datasets and Benchmarks}
Surgical AI has long been supported by a rich ecosystem of public datasets that enable the development and evaluation of data-driven models.
Conventional datasets~\cite{7519080,ayobi2025pixel,wang2022autolaparo,chen2025prostatd} typically provide closed-set categorical labels designed for supervised learning of scene understanding (\eg, instrument recognition) and workflow understanding (\eg, phase recognition).
In parallel, large-scale pre-training datasets~\cite{che2025lemon,wei2025surgbench,ayobi2025pixel,de2025scaling,wu2026unisurg} offer abundant unlabeled or weakly labeled data for representation learning, but their lack of structured annotations limits their effectiveness for fine-grained surgical understanding and multimodal reasoning.
While effective for benchmarking perception and workflow analysis, these datasets remain limited to predefined category labels and fail to capture complex interactions and reasoning over surgical activities, leaving a considerable gap between these datasets and real-world surgical applications. 
Their procedure-centric design also leads to limited cross-procedure generalization and heterogeneous label spaces, posing challenges for unified model training.

With the emergence of multimodal foundation models, recent efforts have shifted toward instruction-following and multi-task datasets tailored to generalizable multimodal surgical modeling. 
Early VQA-style datasets~\cite{seenivasan2022surgical,seenivasan2022surgical,seenivasan2023surgicalgpt,yuan2024advancing} are typically constructed by converting single-task annotations into question–answer pairs, but remain limited in annotation richness and task diversity.
Subsequent works~\cite{liu2025endobench,perez2025surglavi} improve coverage by aggregating multiple open-source datasets, while more recent efforts~\cite{li2025surgpub,low2026surgraw} leverage web-sourced videos to scale multimodal supervision.
In addition, various benchmarks have been developed to evaluate multimodal comprehension in surgical scenarios~\cite{wang2025endochat,choi2025surgmllmbench,rau2025systematic,low2026surgraw}, surgical scene generation~\cite{chen2025far}, interpretability~\cite{cheng2025benchmarking}, and surgical quality assessment~\cite{alapatt2025sages,ban2024concept}.
Despite this progress, existing datasets remain limited in diversity and annotation quality, are biased toward VQA-style conversations, and lack support for dense prediction, spatiotemporal reasoning, planning, and generative tasks. 
Moreover, heterogeneous annotation schemas hinder unified multi-task training, motivating a unified, large-scale dataset for multimodal surgical intelligence.

\subsection{Foundation Models for Surgical Intelligence}
Surgical AI has traditionally relied on task-specific models for instrument/tissue detection and segmentation~\cite{yuan2025systematic,liu2025sam2s}, workflow analysis~\cite{twinanda2016endonet,jin2017sv,jin2022trans} and triplet recognition~\cite{chen2025prostatd,nwoye2020recognition,nwoye2022rendezvous,pei2025instrument}, which are typically trained with procedure-specific supervision and are sensitive to domain shifts across clinical environments. 
With the increasing availability of large-scale surgical video data, recent efforts have moved toward surgical foundation models that learn transferable visual representations via self-supervised~\cite{wang2023foundation,schmidgall2024general,yuan2025learning,yuan2024procedure,yang2026large} or weakly supervised pre-training~\cite{kamabattula2025weakly}. 
These models demonstrate improved robustness and cross-domain generalization, and can be adapted to diverse downstream tasks through lightweight fine-tuning. 
However, such approaches remain primarily perception-centric and lack the ability to support open-ended reasoning, interactive understanding, and high-level decision-making.

Building upon these advances, multimodal foundation models extend foundation modeling by enabling natural-language interaction and reasoning over surgical scenes. 
Early surgical vision-language models focus on VQA-style formulations~\cite{seenivasan2022surgical,PitVQA}, representing surgical elements such as instruments, tissues, and spatial relationships through textual descriptions and generating answers. 
More recently, instruction-tuned surgical multimodal large language models~\cite{wang2024surgical,jin2024surgical,seenivasan2023surgicalgpt,wang2025endochat,wu2026medos,li2024llava,zeng2025surgvlm} and video-level surgical understanding models~\cite{wang2025surgvidlm,WanZip_CholecMamba_MICCAI2025,Yua_HecVL_MICCAI2024} demonstrate strong performance across diverse image- and video-level surgical tasks. 
However, deploying multimodal foundation models in surgical settings remains challenging due to fragmented data resources that lack scale, diversity, unified label spaces, and high-quality reasoning annotations. 
This highlights the need for large-scale, consistently annotated, and unified multimodal datasets for training and evaluation.

\begin{figure}[t]
  \centering
  \includegraphics[width=\linewidth]{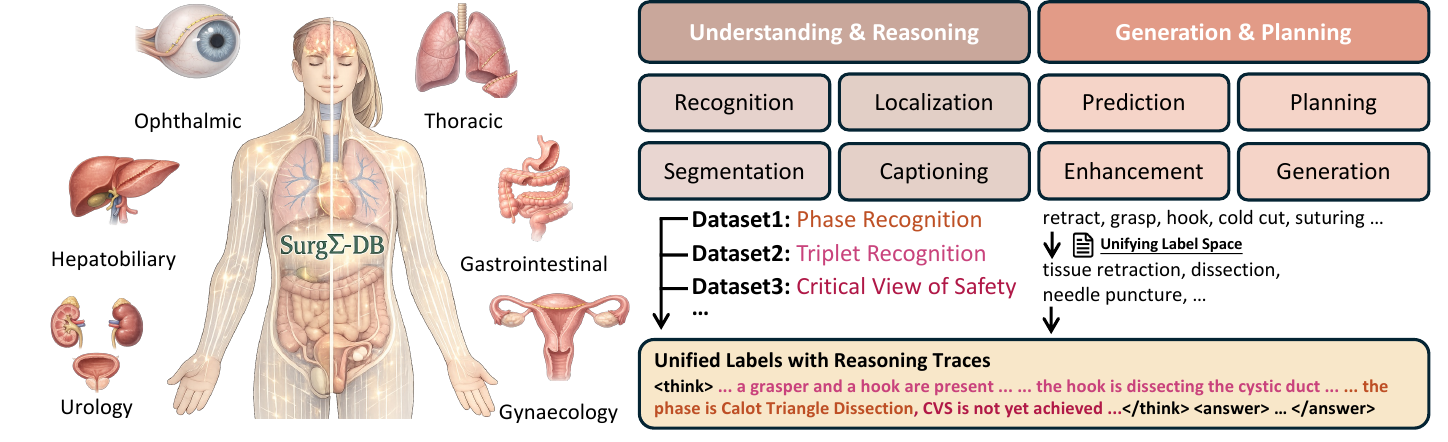}
    \vspace{-2em}
  \caption{Surg$\Sigma$-DB integrates heterogeneous surgical data across 6 clinical specialties into a unified multimodal data foundation. It supports diverse tasks through standardized annotations enriched with hierarchical reasoning traces.}
  \label{fig: dataset overview}
    \vspace{-1em}
\end{figure}

\section{Surg$\Sigma$-DB: A Large-Scale Multimodal Data Foundation for Surgical AI}
Surg$\Sigma$-DB is a large-scale multi-grained dataset constructed for multimodal foundation models in surgical intelligence. 
It contains $\sim$5.98M annotated samples spanning 6 clinical specialties, as shown in Figure~\ref{fig: dataset overview}. 
Surg$\Sigma$-DB integrates rich annotations for both static video frames and video clips, and aligned natural language conversations including instruction–response pairs and reasoning traces.
Data are sourced from both publicly available surgical datasets and curated in-house clinical collections. 
Annotations are produced through a combination of expert human labeling and controlled synthesis pipelines, which further introduce hierarchical reasoning annotations to capture contextual relationships within surgical scenes, ensuring annotation quality, semantic consistency, and scalable coverage.
All subsets are organized under a unified data schema with harmonized label spaces and standardized formats to support multi-task training and benchmarking.

\begin{table*}[t!]
  \centering
  \scriptsize
  \caption{Data sources integrated into Surg$\Sigma$-DB, categorized by clinical specialty and surgical types.}
  \vspace{-1em}
  \label{tab:datasets_by_specialty}
  \begin{tabular}{lllll}
    \toprule
    \textbf{Clinical Specialty} & \textbf{Surgical Type} & \textbf{Data Source} & \textbf{Surgical Platform} & \textbf{Protocol} \\
    \midrule

    \multirow{3}{*}{Gynecologic}
    & \multirow{3}{*}{Hysterectomy}
    & AutoLaparo \cite{wang2022autolaparo} & Non-Robotic & Laparoscopy \\
    & & SurgicalActions160 \cite{schoeffmann2018video} & Non-Robotic & Laparoscopy \\
    & & Web-Collected Data & Both & Laparoscopy \\
    \midrule

    \multirow{2}{*}{Ophthalmic}
    & \multirow{2}{*}{Cataract Surgery}
    & Cataract-1K \cite{ghamsarian2024cataract} & Non-Robotic & OphScope \\
    & & CaDISv2 \cite{luengo20212020} & Non-Robotic & OphScope \\
    \midrule

    \multirow{10}{*}{Hepatobiliary}
    & \multirow{9}{*}{Cholecystectomy}
    & Cholec80 \cite{7519080} & Non-Robotic & Laparoscopy \\
    & & Cholec80-CVS \cite{rios2023cholec80} & Non-Robotic & Laparoscopy \\
    & & CholecInstanceSeg \cite{alabi2025cholecinstanceseg} & Non-Robotic & Laparoscopy \\
    & & CholecT50 \cite{nwoye2022rendezvous} & Non-Robotic & Laparoscopy \\
    & & Endoscapes \cite{mascagni2025endoscapes} & Non-Robotic & Laparoscopy \\
    & & M2CAI16 \cite{twinanda2016endonet} & Non-Robotic & Laparoscopy \\
    & & HeiChole \cite{wagner2023comparative} & Non-Robotic & Laparoscopy \\
    & & Web-Collected Data & Both & Laparoscopy \\

    & & In-House Data & Non-Robotic & Laparoscopy \\
    \cmidrule(lr){2-5}

    & Hepatectomy & In-House Data & Non-Robotic & Laparoscopy \\
    \midrule

    \multirow{10}{*}{Gastrointestinal}
    & \multirow{2}{*}{Gastrectomy}
    & Web-Collected Data & Both & Laparoscopy \\

    & & In-House Data & Non-Robotic & Laparoscopy \\
    \cmidrule(lr){2-5}
    & Gastrojejunostomy & MultiBypass140 \cite{lavanchy2024challenges} & Non-Robotic & Laparoscopy \\
    \cmidrule(lr){2-5}
    & Colonoscopy & SegCol \cite{ju2024segcol} & Non-Robotic & Endoscopy \\
    \cmidrule(lr){2-5}
    & Proctocolectomy & HeiCo \cite{maier2021heidelberg} & Non-Robotic & Laparoscopy \\
    \cmidrule(lr){2-5}
    & Sigmoid Colectomy & HeiCo \cite{maier2021heidelberg} & Non-Robotic & Laparoscopy \\
    \cmidrule(lr){2-5}
    & Rectal Resection & HeiCo \cite{maier2021heidelberg} & Non-Robotic & Laparoscopy \\
    \cmidrule(lr){2-5}
    & Ladd's Procedure & Web-Collected Data & Non-Robotic & Laparoscopy \\
    \cmidrule(lr){2-5}
    & Appendectomy & Web-Collected Data & Non-Robotic & Laparoscopy \\
    \cmidrule(lr){2-5}
    & Rectal Resection/Extirpation & DSAD \cite{carstens2023dresden} & Robotic-Assisted & Laparoscopy \\
    \midrule

    \multirow{10}{*}{Urologic}
    & \multirow{6}{*}{Nephrectomy}
    & EndoVis2017 \cite{allan20192017} & Robotic-Assisted & Laparoscopy \\
    & & EndoVis2018 \cite{allan20202018} & Robotic-Assisted & Laparoscopy \\
    & & Nephrec9 \cite{nakawala2017nephrec9} & Non-Robotic & Laparoscopy \\
    & & SurgT \cite{cartucho2024surgt} & Robotic-Assisted & Laparoscopy \\
    & & Web-Collected Data & Both & Laparoscopy \\

    & & In-House Data & Both & Laparoscopy \\
    \cmidrule(lr){2-5}
    & \multirow{4}{*}{Prostatectomy}
    & GraSP \cite{ayobi2025pixel} & Robotic-Assisted & Laparoscopy \\
    & & SAR-RARP \cite{psychogyios2023sar} & Robotic-Assisted & Laparoscopy \\
    & & MESAD-Real \cite{bawa2021saras} & Robotic-Assisted & Laparoscopy \\
    & & Web-Collected Data & Robotic-Assisted & Laparoscopy \\
    \midrule

    Thoracic
    & Lobectomy
    & Lobectomy Dataset \cite{guo2025surgical} & Robotic-Assisted & Thoracoscopic \\
    
    \bottomrule
  \end{tabular}
\end{table*}

\subsection{Data Curation}
\subsubsection{Multi-Source Surgical Data Collection}
Our goal is to construct a highly diverse surgical data foundation spanning multiple clinical specialties and surgical types. 
Guided by this objective, we collect raw data from a wide range of sources, including publicly available surgical datasets, online surgical videos, and curated in-house clinical collections developed in collaboration with medical partners.
As summarized in Table~\ref{tab:datasets_by_specialty}, we collect 16 surgical types across 6 major clinical specialties (\ie, gynecologic~\cite{wang2022autolaparo,schoeffmann2018video}, ophthalmic~\cite{ghamsarian2024cataract,luengo20212020}, hepatobiliary~\cite{rios2023cholec80,alabi2025cholecinstanceseg,nwoye2022rendezvous,mascagni2025endoscapes,twinanda2016endonet,wagner2023comparative}, gastrointestinal~\cite{lavanchy2024challenges,ju2024segcol,maier2021heidelberg,carstens2023dresden}, urologic~\cite{allan20192017,allan20202018,nakawala2017nephrec9,cartucho2024surgt,ayobi2025pixel,psychogyios2023sar,bawa2021saras} and thoracic~\cite{guo2025surgical} surgeries), encompassing both robotic and manual operations under diverse imaging modalities, including laparoscopic, endoscopic, OphScope, and thoracoscopic settings. 
This breadth ensures substantial variability in anatomy, instrumentation, and workflow dynamics, providing a highly diverse foundation for surgical intelligence modeling.

\subsubsection{Holistic and Multi-Granular Surgical Task Design}
\begin{figure}[t]
  \centering
    \includegraphics[width=\linewidth]{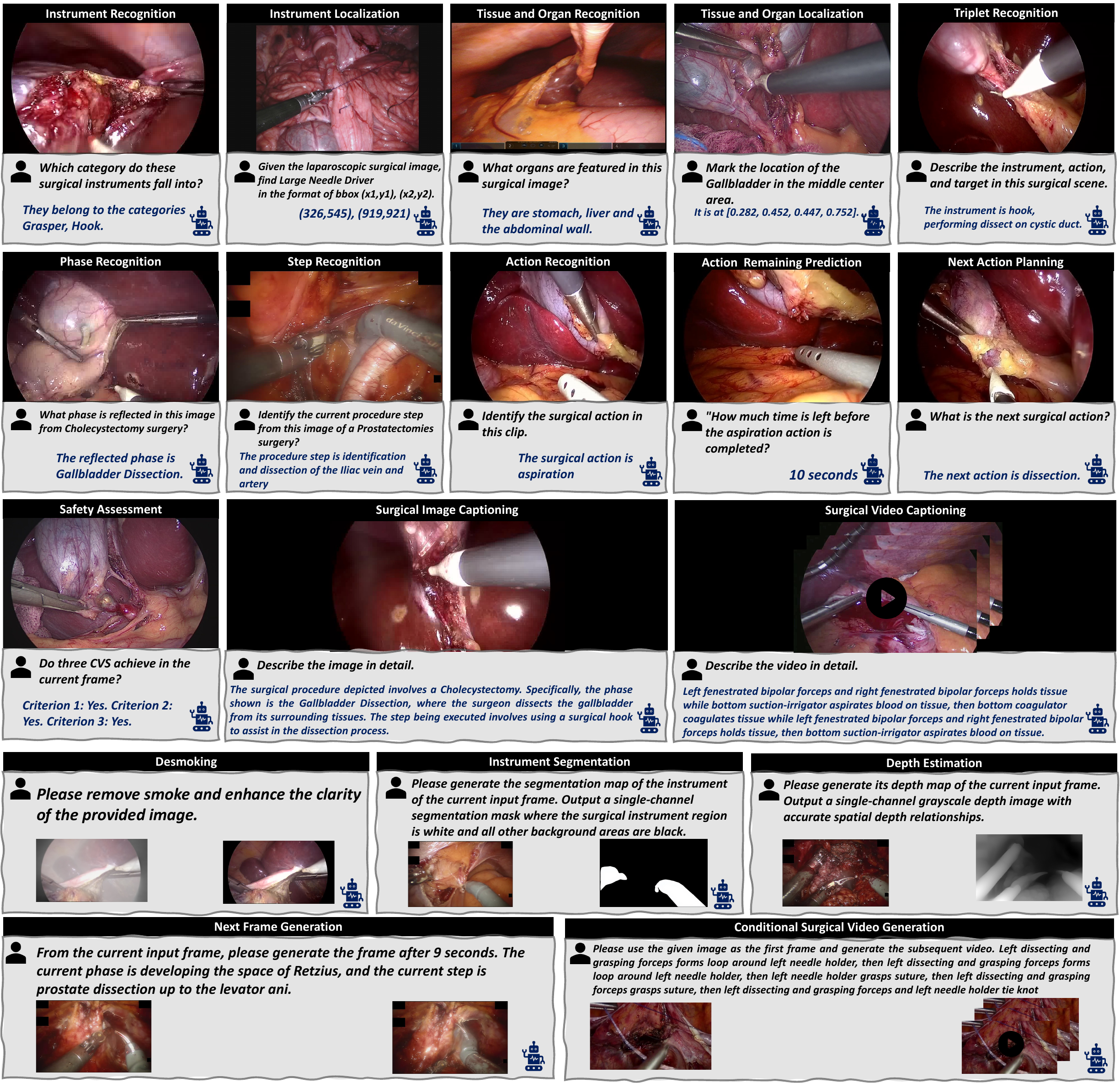}  
  \vspace{-1.5em}
  \caption{Surg$\Sigma$-DB contains diverse multimodal conversations spanning 13 understanding and reasoning tasks as well as 5 planning and generation tasks, supporting a wide range of perception, reasoning, simulation, and decision-oriented capabilities for surgical intelligence.}
  \label{fig: covered tasks}
  \vspace{-1em}
\end{figure}

We design a diverse and multi-granular set of tasks in Surg$\Sigma$-DB to comprehensively cover the key objectives of surgical intelligence, as demonstrated in Figure~\ref{fig: covered tasks}. 
These tasks are organized into two complementary groups: (1) \textbf{Understanding and Reasoning} and (2) \textbf{Planning and Generation}, collectively reflecting the fundamental capabilities required for surgical multimodal foundation models, spanning perception, reasoning, predictive modeling, and controllable content generation.

\textbf{Understanding and Reasoning Tasks.}
These tasks encompass a diverse set of perception and reasoning problems designed to capture multi-granular spatio-temporal understanding of surgical scenes. 
They involve grounding surgical instruments, anatomical structures, and procedural dynamics from both image and video inputs, enabling detailed analysis of tool–tissue interactions and surgical workflows. 
By covering capabilities such as geometric modeling, semantic interpretation, safety verification, and contextual workflow understanding, these tasks reflect both foundational perception challenges and clinically relevant reasoning problems in surgical environments.

\begin{itemize}
\item \textbf{Instrument Recognition:} Identify surgical instruments present in a video frame, serving as a fundamental perceptual capability for understanding tool usage and surgical workflow.

\item \textbf{Instrument Localization:} Predict spatial regions of surgical instruments using either bounding boxes or image patches, enabling precise spatial grounding of tool positions.

\item \textbf{Instrument Segmentation:} Generate pixel-wise masks of surgical instruments to capture fine-grained shapes and tool–tissue occlusion relationships for precise spatial modeling.

\item \textbf{Tissue and Organ Recognition:} Classify visible anatomical entities (\ie, tissues or organs) to establish semantic awareness of operative regions and contextual surgical states.

\item \textbf{Tissue and Organ Localization:} Localize anatomical entities (\ie, tissues and organs) via bounding boxes to provide spatial awareness of anatomical structures during surgery.

\item \textbf{Phase Recognition:} Identify the current surgical phase from either video frames or video clips, recognizing the highest-level procedural stage of the surgical workflow.

\item \textbf{Step Recognition:} Identify finer-grained surgical steps from either video frames or video clips, modeling intermediate-level workflow progression within each phase.

\item \textbf{Action Recognition:} Classify atomic surgical actions from either video frames or video clips, encompassing both basic surgical actions and procedure-specific operations, and capturing the lowest-level dynamics of the surgical workflow.

\item \textbf{Triplet Recognition:} Predict instrument–action–target triplets to explicitly represent structured surgical interactions between instruments and targets (\ie, tissues or other instruments), enabling higher-level understanding and reasoning about surgical activities.

\item \textbf{Depth Estimation:} Infer relative depth maps for surgical video frames, facilitating geometry-aware downstream tasks such as instrument navigation and spatial perception during surgery.

\item \textbf{Safety Assessment:} Determine whether safety-critical anatomical structures have been sufficiently exposed and correctly identified (\eg, critical view of safety~\cite{rios2023cholec80,alapatt2025sages,ban2024concept} in cholecystectomy), ensuring surgical actions such as clipping or transection can be performed safely.

\item \textbf{Surgical Image Captioning:} Generate natural language descriptions of static surgical scenes, summarizing instruments, anatomy, and contextual information visible in inputs.

\item \textbf{Surgical Video Captioning:} Produce temporally coherent textual descriptions of surgical videos, capturing the spatio-temporal evolution of surgical scenes and activities.
\end{itemize}

\textbf{Planning and Generation Tasks.}
Beyond scene comprehension, these tasks focus on predictive modeling and controllable generation that are closely aligned with practical clinical needs. 
They involve forecasting future procedural developments, estimating surgical progress, and generating coherent multimodal descriptions of surgical activities. 
Such tasks provide a basis for applications including intraoperative assistance, workflow anticipation, and simulation-driven learning, highlighting the potential of surgical AI systems to move from passive observation toward proactive support.

\begin{table}[t]
\centering
\caption{Unified annotation format of Surg$\Sigma$-DB.}
  \vspace{-0.5em}
\label{tab: anno format}
\begin{tcolorbox}[colback=gray!5!white,colframe=gray!75!black,boxsep=2pt,left=2pt,right=2pt,top=-5pt,bottom=-5pt]
\begin{lstlisting}[basicstyle=\ttfamily\small, frame=none]
{
  "meta data": 
    {
      "info": "dataset information",
    },
    
  "images": [
      {
        "id": 0,
        "source dataset": "dataset name",
        "source url": "url to data source",
        "clinical specialty": "name of the specialty",
        "surgical type": "name of the surgical procedure",
        "image path": "path to image"
      }
    ],
    
  "videos": [
      {
        "id": 0,
        "source dataset": "dataset name",
        "source url": "url to data source",
        "clinical specialty": "name of the specialty",
        "surgical type": "name of the surgical procedure",
        "video path": "path to video"
      }
    ],

  "annos": [
      {
        "id": 0,
        "videos": [video id],
        "time step": [frame idx],
        "images": [image id],
        "question": "user instructions",
        "thinking": "model thinking response (optional)",
        "answer": "model answer response",
        "dense prediction": ["path to dense prediction"],
        "task type": ["task name"],
        "gt label": ["ground-truth label"],
        "annotator": {"NUS", "CUHK", "SJTU", "NVIDIA"}
      },
    ]
}
\end{lstlisting}
\end{tcolorbox}
\vspace{-1.5em}
\end{table}

\begin{itemize}
\item \textbf{Action Remaining Prediction:} Estimate the remaining duration of the current action, enabling workflow progress assessment and anticipation of upcoming procedural transitions.

\item \textbf{Next Action Planning:} Predict the most probable next surgical action given current visual observation and historical context, enabling anticipation of upcoming workflow transitions.

\item \textbf{Desmoking:} Generate smoke-free surgical video frames from smoke-degraded observations, improving visual clarity and perceptual robustness under real-world imaging conditions.

\item \textbf{Next Frame Prediction:} Generate a future video frame at a specified time horizon to predict the temporal evolution of surgical scenes, supporting workflow anticipation and simulation.

\item \textbf{Conditional Surgical Video Generation:} Generate surgical video clips conditioned on textual instructions and visual context, including text-to-video, image-to-video, and text-guided image-to-video generation, enabling controllable simulation and data augmentation.
\end{itemize}

\begin{table}[t]
\centering
\caption{Three typical annotation samples for MLLM training.}
  \vspace{-0.5em}
\label{tab: instruction format}
\begin{tcolorbox}[colback=gray!5!white,colframe=gray!75!black,boxsep=2pt,left=2pt,right=2pt,top=-5pt,bottom=-5pt]
\begin{lstlisting}[basicstyle=\ttfamily\small, frame=none]
[
  {
    "id": 0,
    "images": ["image path"],
    "conversations": [
      {
        "from": "human",
        "value": "<image>user instruction for image tasks"
      },
      {
        "from": "gpt",
        "value": "<thinking>model thinking response</thinking>\n
                  <answer>model answer response</answer>"
      }
    ]
  },

  {
    "id": 1,
    "videos": ["video path"],
    "conversations": [
      {
        "from": "human",
        "value": "<image>user instruction for video tasks"
      },
      {
        "from": "gpt",
        "value": "<thinking>model thinking response</thinking>\n
                  <answer>model answer response</answer>"
      }
    ]
  },

  {
    "id": 2,
    "images": ["image path (input)", "image path (prediction)"],
    "conversations": [
      {
        "from": "human",
        "value": "<image>user instruction for dense prediction tasks"
      },
      {
        "from": "gpt",
        "value": "<image>"
      }
    ]
  }
]
\end{lstlisting}
\end{tcolorbox}
\vspace{-1.25em}
\end{table}

\subsubsection{From Heterogeneous Labels to Unified Annotations}
Surgical datasets collected from diverse sources often exhibit inconsistent terminology, mismatched category definitions, and varying annotation formats, especially for fine-grained units such as atomic surgical actions. 
These discrepancies arise from divergent naming conventions, variations in semantic granularity, and heterogeneous representation forms (\eg, categorical labels, dense masks, or question–answer pairs with multi-step reasoning), leading to semantic drift and instability in large-scale joint training.
To address this, we reorganize heterogeneous labels into a unified framework that aims to standardize fine-grained semantic definitions, reduce cross-dataset inconsistencies, and support scalable, interoperable training.

Taking action recognition as a representative example, we consolidate diverse atomic actions into a unified taxonomy of ten basic surgical actions with explicit semantic boundaries and inclusion criteria. 
Each action is grounded in clinically interpretable descriptions and aligned with related attributes (\eg, instruments and target tissues), forming a coherent and structured label space that supports consistent supervision across heterogeneous surgical data.

Beyond semantic normalization, we unify heterogeneous annotation formats under a consistent structural schema. 
Different annotation types and tasks are reorganized into a standardized representation that aligns multi-granular supervision across image- and video-level data, as illustrated in  Table~\ref{tab: anno format}. 
This unified structure can be readily converted into training-ready multimodal format (see Table~\ref{tab: instruction format}) compatible with large language models, enabling direct integration into foundation model pipelines. 
Such standardized annotation format ensures structural consistency, facilitates large-scale joint training, and supports future dataset scaling and extensibility.

\begin{figure}[t]
  \centering
  \includegraphics[width=\linewidth]{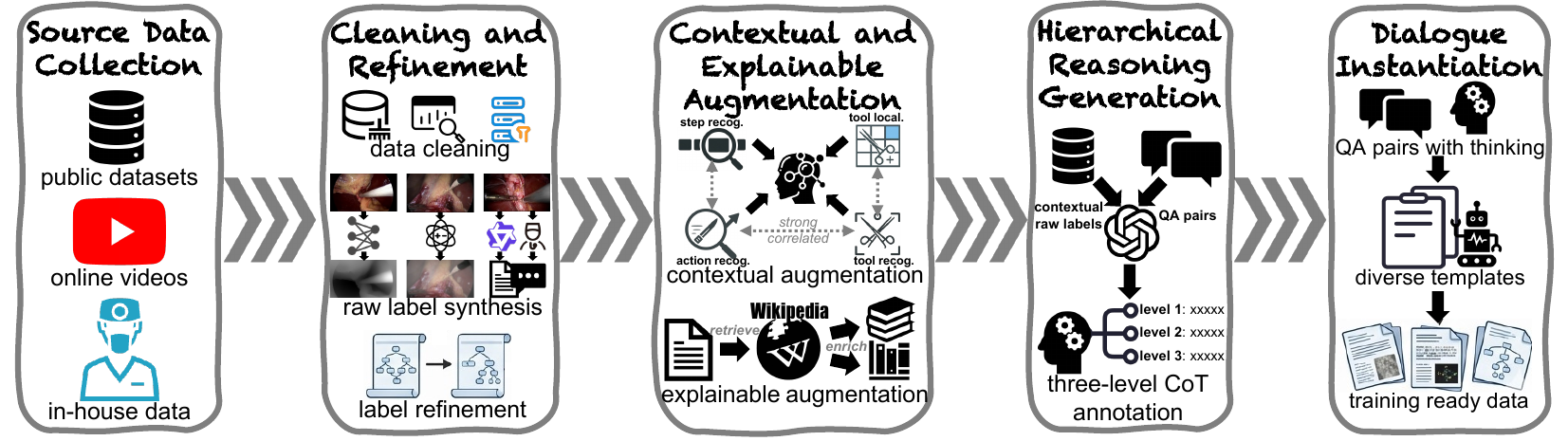}\
  \vspace{-1.5em}
  \caption{Overview of the data curation and annotation pipeline for Surg$\Sigma$-DB.}
  \label{fig: annotation pipeline}
  \vspace{-1em}
\end{figure}

\subsubsection{Semi-Automated Annotation Pipeline}

\textbf{Data Pre-Processing and Label Refinement.}
Following aforementioned unified annotation standards, we systematically refine raw labels from heterogeneous sources to ensure semantic and structural consistency. 
Coarse categories, institution-specific shorthand, and inconsistent terminology are replaced with explicit, context-complete descriptions and normalized under standardized medical vocabularies. 
For instance, ambiguous placeholders (\eg, ``other'') are reformulated into precise statements, and synonymous anatomical terms (\eg, ``Calot’s triangle'' \textit{vs.} ``Hepatocystic triangle'') are consolidated into canonical forms. 
This standard-driven refinement reduces cross-dataset ambiguity and improves stability in large-scale joint training.

For the raw annotations that are already provided by the original source datasets, we directly adopt the official labels to preserve dataset fidelity.
For incomplete textual annotations (\eg, image and video captioning), we leverage Qwen3-VL-235B~\cite{bai2025qwen3} to generate enriched and context-consistent descriptions.
For missing dense predictions, such as smoke masks, segmentation maps, and depth annotations, we employ off-the-shelf methods~\cite{he2010single,ravisam,yang2024depth} to automatically generate the corresponding information.
For some noise-prone labels (\eg, temporal boundaries of surgical actions), we perform manual verification and label refinement to ensure precise and reliable annotations.

\textbf{Contextual and Explainable Augmentation.}
Surgical attributes such as phase, step, instrument, and action are inherently interdependent. 
To enhance structural learning, we optionally consolidate correlated attributes into unified prompts that explicitly encode hierarchical and relational dependencies (\eg, instrument–action pairs), transforming isolated labels into structured multi-attribute supervision. 
In addition, to strengthen vision-language alignment, we augment original categorical labels with knowledge-grounded descriptions derived from authoritative medical sources (\eg, Wikipedia), linking visual evidence to surgical intent and anatomical context. 
Together, these contextual and explanation-aware augmentations promote compositional reasoning, improve fine-grained alignment, enhance robustness across diverse scenes, and increase interpretability for real-world deployment.

\textbf{Hierarchical Reasoning Trajectory Generation.}
To explicitly align visual evidence with structured reasoning processes, a three-level chain-of-thought~\cite{wei2022chain} annotation strategy is constructed to decompose surgical inference into perceptual grounding, relational understanding, and contextual reasoning, as shown in Figure~\ref{fig: annotation pipeline}. 
Specifically, the first level emphasizes describing fundamental visual elements present in the scene, focusing on the appearance and spatial properties of tools and tissues. 
The second level focuses on characterizing interactions among different elements, capturing how tools, tissues, and actions relate to and influence one another through contact, motion, and functional associations. 
The third level abstracts perceptual and relational cues into high-level procedural inference, enabling structured reasoning about the global surgical context.

These structured reasoning trajectories are synthesized using GPT-5.1~\cite{singh2025openai} conditioned on all verified raw labels (from existing datasets or annotated by ourselves) under explicit forward-reasoning constraints, ensuring that intermediate steps remain grounded in observable visual evidence and preventing hallucinated interactions beyond the annotated scene.
Consequently, the hierarchical CoT annotations transform flat supervision into structured reasoning guidance, promoting compositional inference, mitigating shortcut learning, and enhancing interpretability.

\textbf{Conversational Diversity Expansion.}
To prepare multimodal foundation model training data, structured annotations are transformed into diverse conversational formats. 
Rigid template design may lead to overfitting in MLLMs, as the model can memorize superficial linguistic patterns rather than learning underlying visual–semantic alignment. 
To mitigate this issue, we construct 100–200 dialogue templates with controlled variations in phrasing, ordering, and information density. 
This strategy increases linguistic diversity while preserving logical consistency, thereby improving instruction-following robustness and cross-task generalization.

\begin{figure}[t]
  \centering
  \begin{subfigure}[t]{0.3\linewidth}
  \centering
  \includegraphics[width=\linewidth]{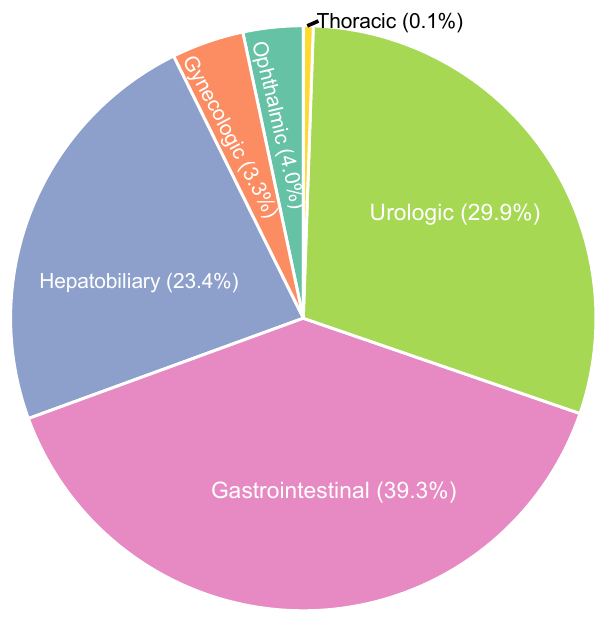}
  \caption{ratio of clinical specialties}
  \label{fig: clinical specialties}
  \end{subfigure}
  \hfill
  \begin{subfigure}[t]{0.3\linewidth}
  \centering
  \includegraphics[width=\linewidth]{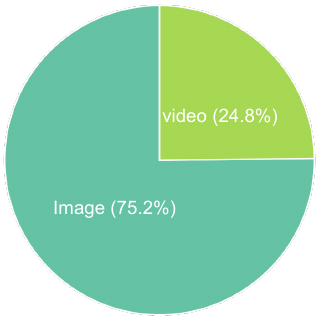}
  \caption{ratio of visual input modalities}
  \end{subfigure}
  \hfill
  \begin{subfigure}[t]{0.3\linewidth}
  \centering
  \includegraphics[width=\linewidth]{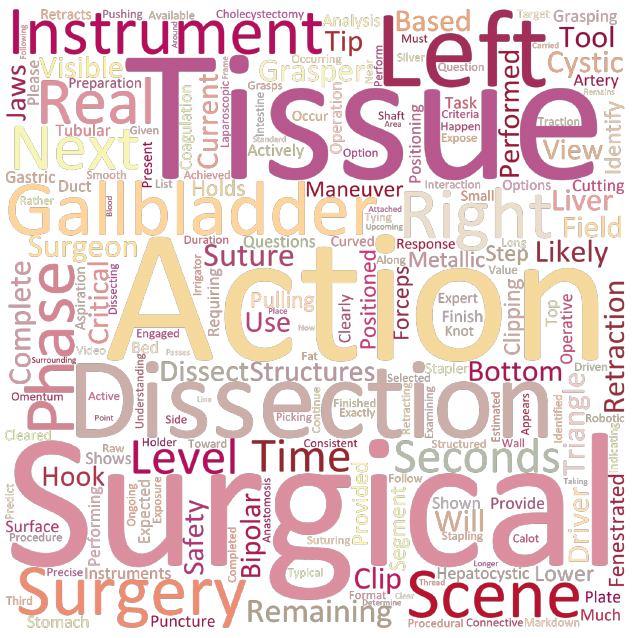}
  \caption{word cloud visualization}
  \end{subfigure}\\
     \vspace{0.5em}
  \begin{subfigure}[t]{0.495\linewidth}
  \centering
  \includegraphics[width=\linewidth]{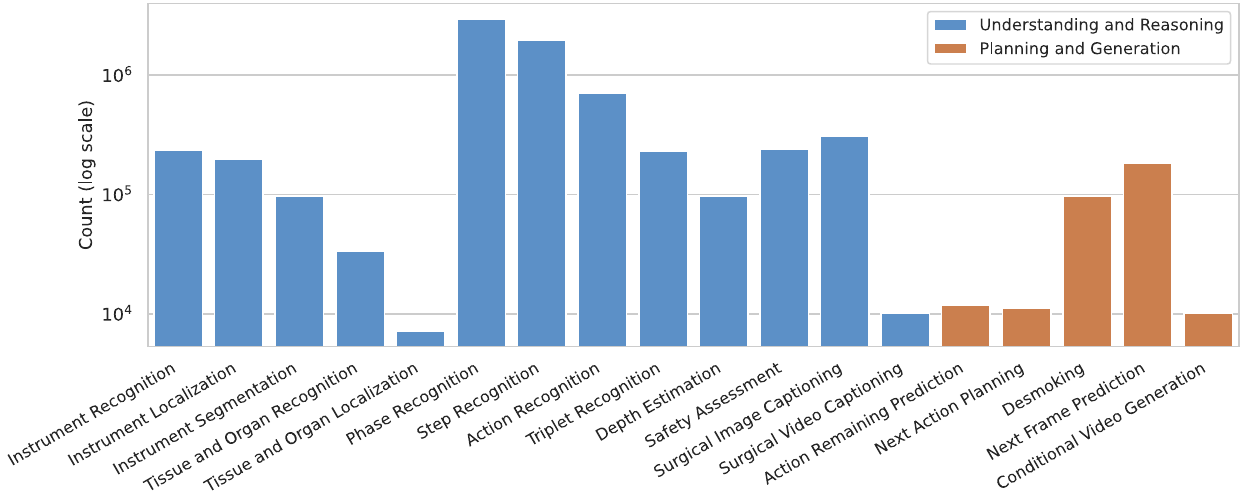}
   \vspace{-1.5em}
  \caption{distribution of all covered tasks}
  \label{fig: task dist}
  \end{subfigure}
  \hfill
  \begin{subfigure}[t]{0.495\linewidth}
  \centering
  \includegraphics[width=\linewidth]{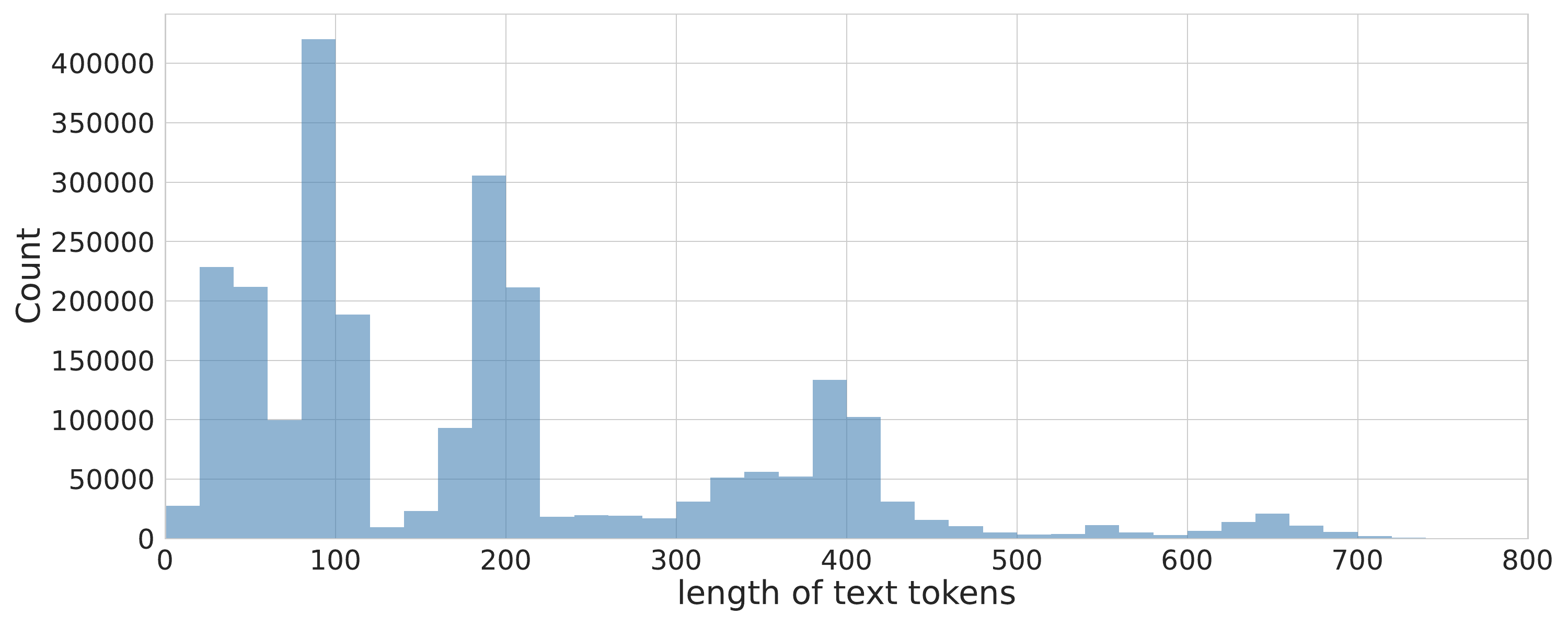}
   \vspace{-1.5em}
  \caption{distribution of text token lengths}
  \label{fig: text length dist}
  \end{subfigure}
  \caption{Statistical analysis of the constructed Surg$\Sigma$-DB.}
  \label{fig: dataset analysis}
  \end{figure}

\subsection{Dataset Analysis and Discussions}
\subsubsection{Data Statistics}
Surg$\Sigma$-DB contains 5.98M multimodal conversations spanning both image- and video-based surgical tasks, extracted from 1.59K unique surgical video sources. 
Among them, 4.49M conversations are associated with images, while 1.48M correspond to video clips, forming a large-scale multimodal corpus for surgical intelligence.
In total, Surg$\Sigma$-DB comprises 471.29M text tokens, providing rich linguistic corpus for multimodal learning. 
It covers 6 clinical specialties and 16 surgical procedure types, capturing diverse surgical scenes and procedural contexts across a wide range of operative domains, as illustrated in Figure~\ref{fig: clinical specialties}.

In terms of visual data, Surg$\Sigma$-DB comprises 1.58M unique images paired with 4.21M image-based conversations. 
This large-scale image corpus supports a wide range of perception and reasoning tasks, including recognition, localization, segmentation, and captioning.
For the video component, Surg$\Sigma$-DB contains 1.35M video clips paired with 1.45M conversations. 
These clips are typically short segments capturing fine-grained surgical activities and procedural context, supporting tasks such as action recognition and conditional surgical video generation.

The scale, diversity and multimodal richness of Surg$\Sigma$-DB make it a strong foundation for training multimodal foundation models for surgical intelligence.
As shown in Figure~\ref{fig: task dist}, Surg$\Sigma$-DB exhibits broad coverage across diverse tasks, while Figure~\ref{fig: text length dist} reveals a wide distribution of text token lengths, indicating rich linguistic diversity spanning both concise perception queries and complex multi-step reasoning.

\subsubsection{Dataset License and Accessibility}
Surg$\Sigma$-DB is licensed under the Creative Commons Attribution-NonCommercial-ShareAlike 4.0 International License (CC BY-NC-SA 4.0). 
The license applies to all annotations to which we have directly contributed. 
Surg$\Sigma$-DB also incorporates surgical videos and images sourced from pre-existing collections. 
For these data, the original licensing terms are respected and remain applicable.

Our first public release, Surg$\Sigma$-DB v0.1, will be made publicly accessible through the official project page, where users can obtain the annotation files, metadata, and documentation necessary to reproduce the structure of the dataset. 
Surg$\Sigma$-DB is intended for non-commercial research purposes, and users are expected to properly cite both Surg$\Sigma$-DB and the original datasets in any resulting publications.

\section{Advanced Foundation Models Built upon Surg$\Sigma$-DB}
Building upon Surg$\Sigma$-DB, a spectrum of foundation models are developed, which cover complementary dimensions of surgical intelligence, from action-centric understanding and multimodal surgical scene understanding to structured reasoning and embodied policy learning.
In the following, we briefly describe each model, highlighting its model and training designs, as well as empirical findings.

\begin{figure*}[t]  
	\centering
	\includegraphics[width=\linewidth]{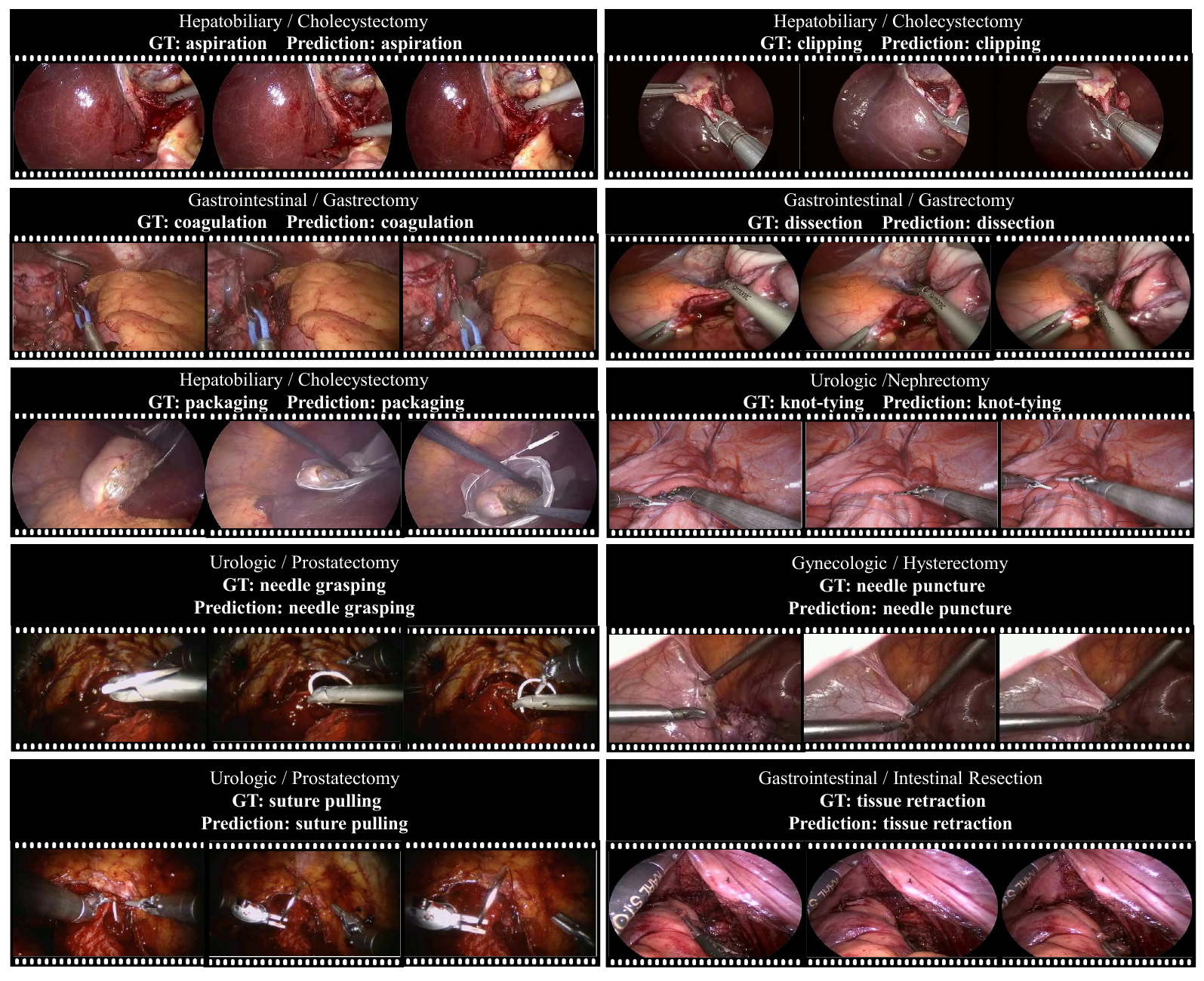}
    \vspace{-1em}
	\caption{Qualitative visualization of BSA foundation model predictions across diverse surgical procedures.}
    \label{fig: BSA results}
  \vspace{-1em}
\end{figure*}

\subsection{BSA: A Cross-Specialty Foundation Model for Basic Surgical Action Recognition}
BSA~\cite{xu2026bsa} is a cross-specialty foundation model for recognizing basic surgical actions as a shared semantic unit across diverse procedures. 
Rather than modeling each procedure in isolation, BSA treats surgical workflow as compositions of reusable primitive actions (\eg, dissection, coagulation, clipping, knot-tying), enabling a unified representation that transfers across anatomical sites, institutions, and recording conditions. 
Given short surgical video clips as input, it outputs probability distributions over ten predefined surgical action categories.
Specifically, BSA builds upon a Video Transformer backbone~\cite{dosovitskiy2021imageworth16x16words} with two key design considerations for surgical video analysis: 
(1) sequential temporal and spatial attention mechanisms are utilized to effectively capture spatiotemporal dependencies inherent in surgical procedures, such as instrument movements and tissue interactions;
(2) a dual-head prediction module is specifically designed to address the class imbalance issue.
With video-action samples in Surg$\Sigma$-DB, the training pipeline utilized standard video preprocessing with temporal downsampling and uniform frame sampling to balance computational efficiency with visual information preservation.
The Evidential loss~\cite{sensoy2018evidential} is employed to encourage well-calibrated uncertainty estimates, preventing both overconfident and excessively uncertain predictions.
Please refer to BSA~\cite{xu2026bsa} for more implementation details.

Experimental results show that BSA learns stable and transferable representations of basic surgical actions across heterogeneous procedures, institutions, and imaging conditions (see Figure~\ref{fig: BSA results}). 
Beyond recognition, BSA provides structured and uncertainty-aware action semantics that naturally support downstream applications, including surgical skill assessment and surgical action planning.
These findings indicate that BSA functions not only as a standalone recognition model, but also as a foundational perception module that bridges low-level visual understanding and higher-level reasoning systems and embodied policy-learning pipelines.
From a data-centric perspective, BSA operationalizes three principles for scalable surgical intelligence.
First, ontology-first supervision: defining a compact, clinically meaningful action vocabulary improves semantic consistency across datasets and specialties. Second, cross-specialty alignment: harmonized labels and standardized preprocessing reduce dataset-specific shortcuts and encourage representations that generalize beyond a single procedure type. Third, uncertainty-aware recognition: modeling confidence is essential for safe handoff to reasoning and control modules in high-stakes clinical settings. 
Together with SurgVLM (large-scale perception and reasoning alignment), Surg-R1 (hierarchical structured surgical reasoning), and Cosmos-H-Surgical (world-model-driven action-data synthesis for robot policy learning), BSA forms the front-end perceptual anchor of a unified pipeline toward embodied surgical intelligence: perception $\rightarrow$ reasoning $\rightarrow$ action.
This systems view suggests that scalable surgical AI depends not on isolated model gains, but on tight co-design of action ontology, multimodal reasoning, and physically grounded world models.

\begin{figure*}[t]  
	\centering
	\includegraphics[width=1\linewidth]{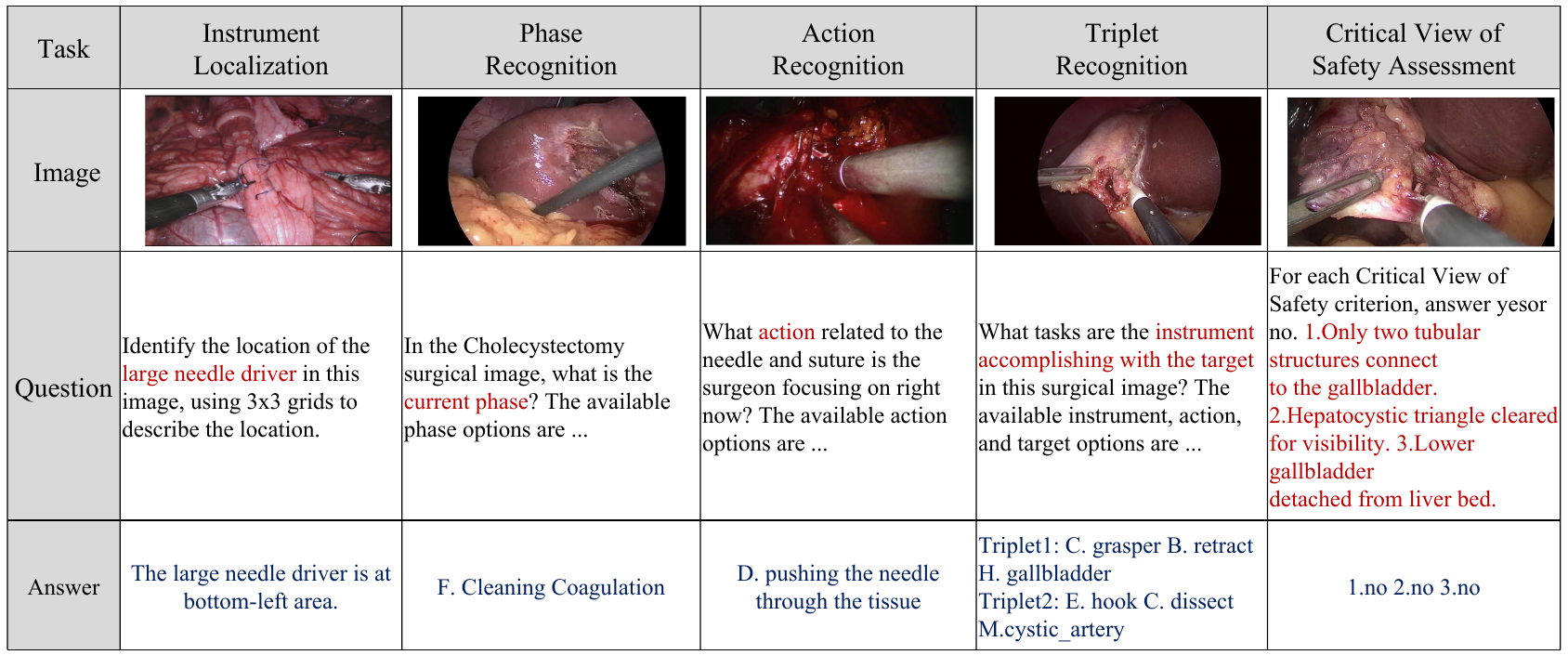}
  \vspace{-1.5em}
	\caption{Qualitative results of SurgVLM-72B including five typical examples from visual perception to temporal analysis to safety reasoning. For triplet recognition, the output format is triplet list with <Instrument,Verb,Target>.}
    \label{Figure:4}
  \vspace{-1em}
\end{figure*}

\subsection{SurgVLM: A Multimodal Foundation Model for Surgical Intelligence}

Surgical intelligence presents unique challenges, requiring surgical visual perception, temporal analysis, and reasoning. 
Existing general-purpose vision-language models fail to address these needs due to insufficient domain-specific supervision and the lack of a large-scale high-quality surgical database.
General-purpose vision-language models, trained predominantly on natural images and text, often exhibit inefficiency by generating excessive, clinically irrelevant outputs. Additionally, their outputs tend to be ambiguous, presenting multiple plausible scenarios rather than definitive, medically meaningful answers. Such ambiguity and verbosity undermine their alignment with surgeons' professional standards and real-world clinical requirements, significantly limiting their reliability and applicability in surgical practice.
To address this challenge, SurgVLM is built upon Surg$\Sigma$-DB, adapted to ten surgical tasks through a unified sequence-to-sequence formulation optimized with a single autoregressive language modeling loss.
To enhance generalization and mitigate biases, it develops an effective database construction pipeline, including data cleaning and refinement, cross-task correlation enrichment, explainable answer generation, and conversational diversity expansion.
As a multimodal foundation model available in multiple scales (\ie, 7B, 32B, and 72B), SurgVLM is designed to support a wide range of surgical understanding tasks, spanning both spatial and temporal analysis of surgical scenes, covering capabilities from visual perception to high-level reasoning. 
To build an effective training pipeline, SurgVLM models follow Qwen2.5-VL~\cite{bai2025qwen25vltechnicalreport} architecture, consisting of a vision encoder, a projector, and a LLM decoder. 
The vision encoder is a transformer-based image backbone that processes images and video frames at their native resolution, while the LLM serves as the decoder for generating outputs.
Please refer to SurgVLM~\cite{zeng2025surgvlm} for more implementation details.

To systematically evaluate multimodal surgical intelligence, SurgVLM is evaluated on SurgVLM-Bench, a comprehensive benchmark designed to assess vision-language models across clinically relevant dimensions of surgical understanding. SurgVLM-Bench integrates six widely used surgical datasets spanning three hierarchical levels of task complexity: visual perception, temporal workflow analysis, and safety reasoning. These task categories reflect increasing contextual and temporal dependency and align with the requirements of real-world surgical assistance.
The qualitative results with typical examples shown in Figure \ref{Figure:4} generated by SurgVLM-72B, including instrument localization, phase recognition, action recognition, triplet recognition, and CVS assessment.
The experiments of SurgVLM indicate that fine-tuning general VLMs on Surg$\Sigma$-DB provides an efficient and reliable pathway for surgical adaptation with following two important insights:
(1) It indicates one of core problems in surgical foundation modeling lies in balancing diversity across surgical types with the exploitation of shared cross-procedure structure. Related procedures often exhibit substantial overlap in anatomical appearance, tissue characteristics, and instrument-associated visual cues, and joint training on multiple categories in Surg$\Sigma$-DB allows the model to leverage these synergies to learn richer and more transferable representations. At the same time, substantial variation remains across procedures in anatomy, workflow, and visual distribution; accordingly, Surg$\Sigma$-DB incorporates the surgical type as explicit contextual information during instruction tuning, reducing ambiguity and improving procedure-specific conditioning. 
(2) It supports a hierarchical view of surgical intelligence in which low-level perception, temporal understanding, and high-level reasoning are tightly coupled. Accurate recognition of instruments and tissues provides the basis for phase and action understanding, while robust temporal modeling underpins reliable intraoperative reasoning. This structure motivates multi-task co-training and a curriculum that progresses from perception to temporal analysis and finally to reasoning, improving data efficiency, convergence behavior, and robustness across the full surgical workflow.

\subsection{Surg-R1: A Reasoning-Enhanced Model for Surgical Scene Understanding}

\begin{figure*}[t]
\centering
\includegraphics[width=\textwidth]{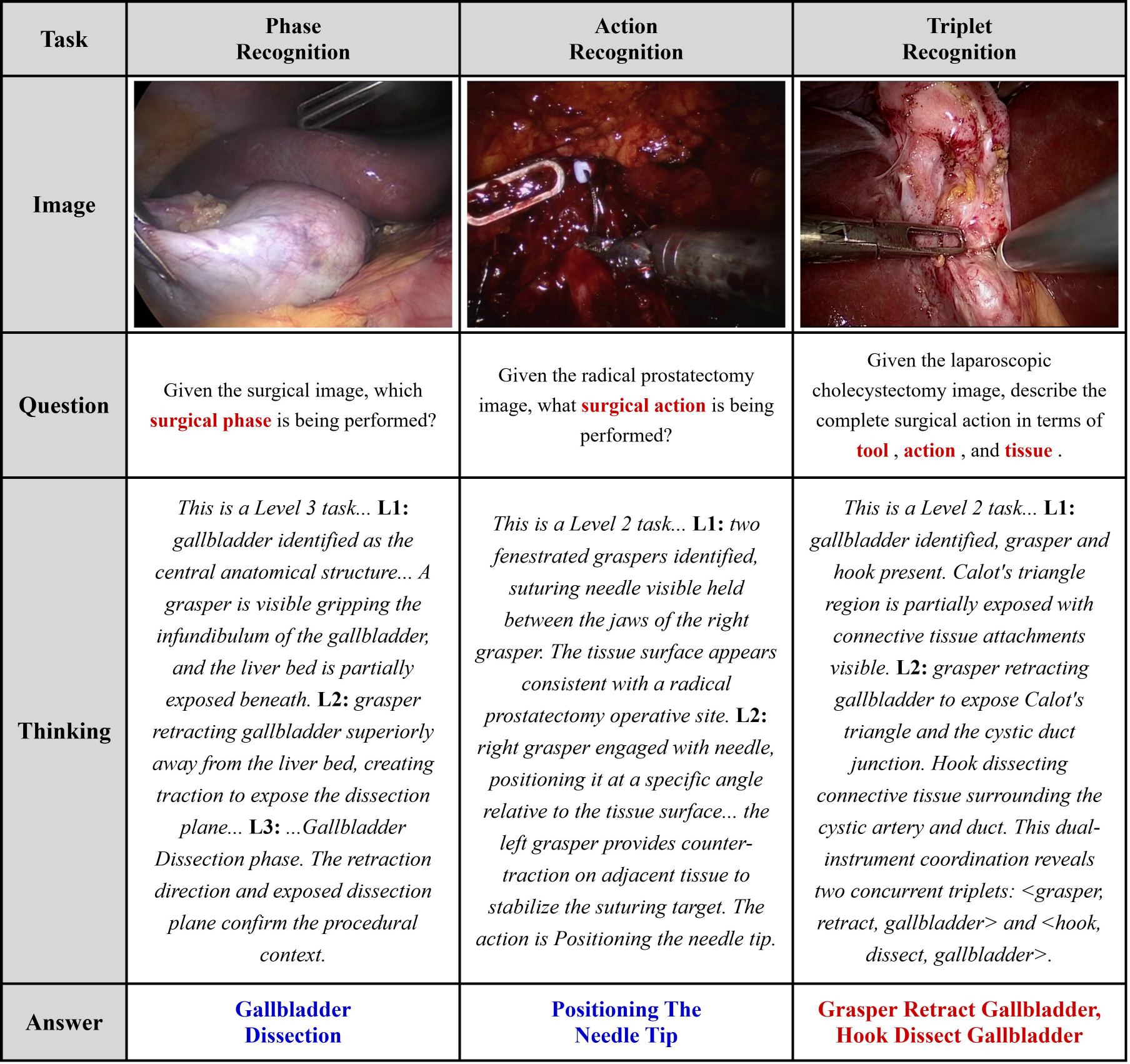}
  \vspace{-1em}
\caption{Qualitative results of Surg-R1-7B across three representative surgical tasks. Each column shows the model's structured multi-level reasoning chain, progressing from visual identification (Level~1) through tool-tissue interaction analysis (Level~2) to procedural understanding (Level~3). \textbf{L$x$} denotes reasoning Level~$x$. Reasoning traces are abbreviated with ellipses (\ldots) for brevity.}
  \vspace{-1em}
\label{fig: Surg-R1 results}
\end{figure*}

Surg-R1~\cite{jiang2026surgr1} is a reasoning-enhanced multimodal foundation model for surgical scene understanding with hierarchical chain-of-thought~\cite{wei2022chain} reasoning capabilities.
Built upon the reasoning-annotated data within Surg$\Sigma$-DB, Surg-R1 interprets complex surgical scenes through a structured three-level reasoning hierarchy: (1) perceptual grounding for instrument and tissue identification, (2) relational understanding for tool-tissue-action interactions, and (3) contextual reasoning for phase recognition and safety assessment. 
Initialized with Qwen2.5-VL-7B~\cite{bai2025qwen25vltechnicalreport}, Surg-R1 is trained through a comprehensive four-stage pipeline: 
First, supervised fine-tuning is performed to establish foundational vision–language alignment using question-answer pairs without reasoning.
Secondly, structured reasoning priors are introduced through cold-start fine-tuning on reasoning trajectories synthesized under surgery-aware constraints that encode structured domain knowledge.
Then, reasoning capability is further refined via reinforcement learning using Group Relative Policy Optimization~\cite{shao2024deepseekmath}.
Finally, an iterative refinement stage combines rejection sampling for correctly predicted samples and teacher-guided knowledge distillation for hard cases, progressively improving reasoning generalization capabilities beyond initial training data.
Please refer to Surg-R1~\cite{jiang2026surgr1} for more implementation details.

Surg-R1 is evaluated on thirteen datasets spanning six core surgical AI tasks, with seven public benchmarks and six multi-center external validation sets from five institutions, against proprietary reasoning models (GPT-5.1, Gemini 3.0 Pro), open-source generalist VLMs, and surgical-domain baselines. As shown in Figure~\ref{fig: Surg-R1 results}, Surg-R1 produces structured multi-level reasoning chains that ground predictions in visual evidence. Surg-R1 achieves state-of-the-art performance across both settings, with the largest gains on compositional tasks. On CholecT50 triplet recognition, for example, Surg-R1 attains 51.69\% accuracy versus 6.77\% for GPT-5.1 and 8.01\% for Qwen2.5-VL-7B-Surg. On multi-center external data it achieves an average arena score of 60.0\%, compared with 44.9\% for the leading surgical baseline. From a data-centric perspective, the consistent failure of general-purpose chain-of-thought reasoning on surgical compositional tasks, even in frontier models such as GPT-5.1, highlights the necessity of domain-specific structural priors. Surg$\Sigma$-DB's multi-granular annotation taxonomy provides the hierarchical scaffolding that makes effective surgical reasoning possible, and its structured instrument, tissue, and action vocabularies anchor the CoT synthesis pipeline in visual observations, suppressing the hallucination artifacts that arise when models reverse-engineer explanations from labels. The three-level reasoning structure also brings practical advantages beyond training. Heterogeneous downstream systems can selectively consume only the granularity level relevant to their task. Skill assessment modules analyze Level~2 interaction patterns, workflow management systems leverage Level~3 phase predictions, and safety monitors extract CVS criteria, all without post-processing unstructured natural-language output. The structured hierarchy further enables automatic cross-level consistency checking and precise fault localization when predictions are wrong. Finally, every inference produces a hierarchically labeled reasoning trace that, after clinical review, can be ingested as new supervision, forming a data flywheel that continuously lowers annotation cost and grows the structured reasoning corpus in Surg$\Sigma$-DB.

\subsection{Cosmos-H-Surgical: A Surgical World Model for Scalable Robot Policy Learning}
Cosmos-H-Surgical~\cite{he2025surgworld} is a surgical world model and data-augmentation pipeline designed to bridge the gap between abundant unlabeled surgical video and the scarce paired video–kinematics data required for surgical robot vision–language–action (VLA) policy training.
Based on Surg$\Sigma$-DB, a surgery-focused captioned dataset is curated with expert-authored action descriptions aligned to short surgical clips.
The generative backbone of Cosmos-H-Surgical is based on a state-of-the-art video world model~\cite{ali2025world}. 
During training, Cosmos-H-Surgical learns to model surgical scene appearance and spatiotemporal dynamics under typical surgical imaging nuisances (specular highlights, occlusions, constrained tool motion), and condition generation on fine-grained text descriptions so that synthesized videos preserve actionable affordances and tool–tissue interactions required for downstream policy learning. 
An inverse-dynamics model (IDM) is trained on limited real paired demonstrations (when available) and then applied to synthetic video to recover pseudo kinematics (approximate action/robot-state sequences), thus producing large-scale synthetic (video, pseudo-kinematics, text) triples suitable for supervised VLA or imitation-style policy learning. 
Cosmos-H-Surgical therefore turns unlabeled surgical video corpora into paired training data at scale, enabling standard VLA optimization without requiring dense manual robot-state annotation.
Please refer to Cosmos-H-Surgical~\cite{he2025surgworld} for more implementation details.

\begin{figure}[t]
  \centering
  \includegraphics[width=1\linewidth]{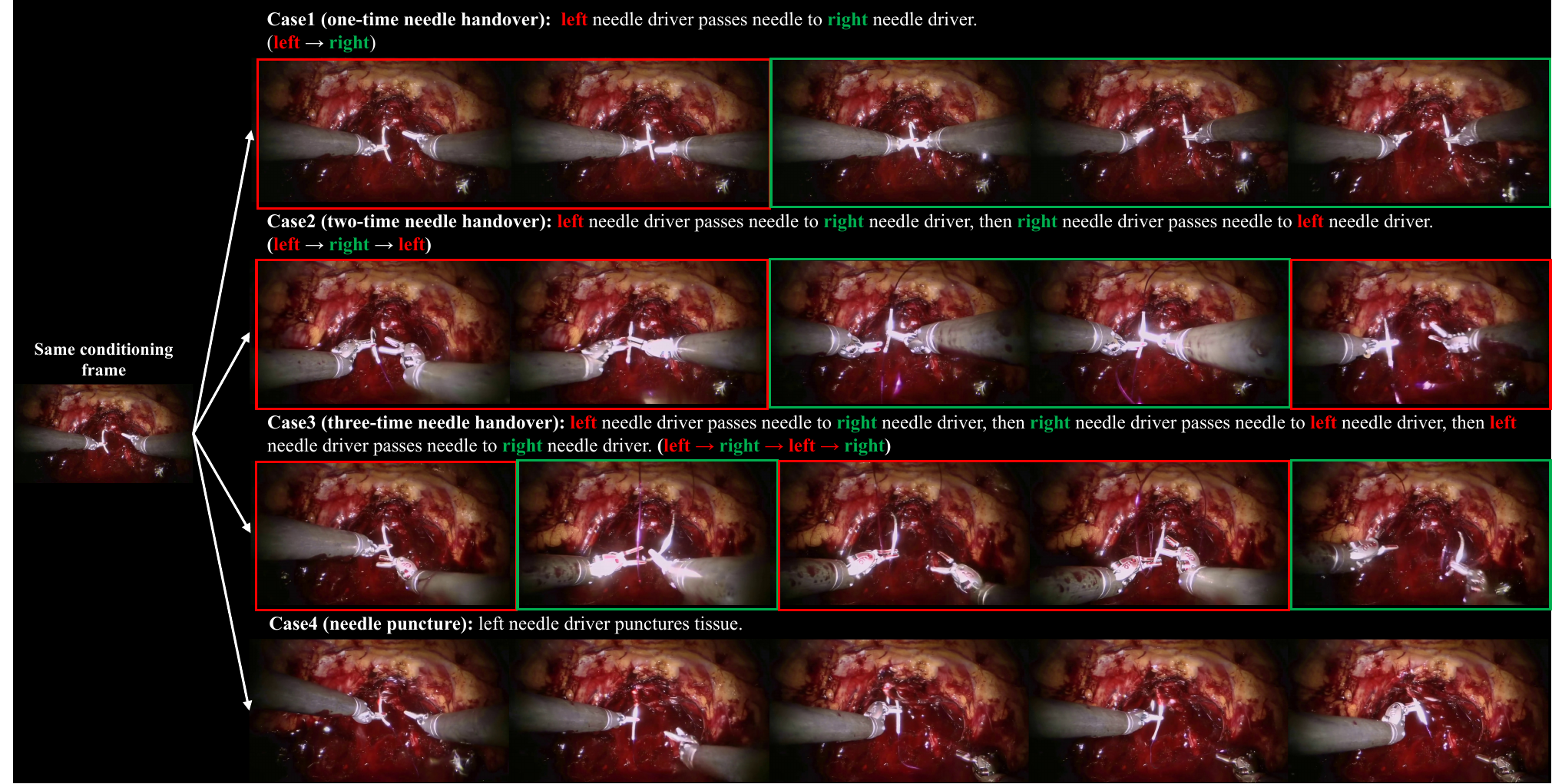}
  \vspace{-1em}  
  \caption{Cosmos-H-Surgical results: New behavior generalization via strong text–video alignment. Given the same conditioning frame, our surgical world model generates distinct video rollouts corresponding to four task prompts: (1) one-time needle handover, (2) two-time needle handover, (3) three-time needle handover, and (4) needle puncture. }
\vspace{-1em}
  \label{fig: Cosmos-H-Surgical results}
\end{figure}

Experimental results show that Cosmos-H-Surgical–augmented policies significantly outperform those trained solely on limited real demonstrations, achieving higher task success rates and improved sample efficiency. Although IDM-generated pseudo-kinematics are imperfect, they provide sufficiently informative supervision when paired with realistic world-model synthesis. The combination of generative diversity and structured inverse inference enables effective scaling of embodied training data. From a data-centric perspective, Cosmos-H-Surgical highlights three key insights. First, domain-specific data curation is essential. Generic video generation is insufficient for surgical scenarios; the SATA dataset, curated from BSA with structured action annotations, provides the semantic grounding required to align generation with physically meaningful surgical actions. Second, forward physical consistency matters. World models must capture constrained tool kinematics and tissue deformation; otherwise, synthetic demonstrations can introduce policy bias. Third, hybrid supervision is most effective. While synthetic augmentation reduces reliance on real demonstrations, the best performance arises from mixed training (synthetic + limited real data), where real data anchors policies within true robot dynamics. Together with SurgVLM (large-scale perception and reasoning alignment) and Surg-R1 (hierarchical structured reasoning with reinforcement refinement), Cosmos-H-Surgical completes the pipeline toward embodied surgical intelligence: perception → reasoning → action. These results suggest that scalable surgical AI requires not only multimodal foundation models, but also structured world models that translate visual understanding into physical control.

\section{Limitations and Future Work}
We aim to progressively achieve comprehensive multi-task annotation coverage across all surgical scenes in Surg$\Sigma$-DB. 
Although its current v0.1 release spans diverse understanding, reasoning, planning, and generation tasks, full-spectrum supervision has not yet been uniformly established for every sample. 
While certain subsets already provide comprehensive multi-task and reasoning-level annotations, other samples remain limited to task-specific supervision (\eg, perception-level labels), and structured reasoning annotations are not consistently available across all conversations.

This imbalance largely stems from the intrinsic complexity and high cost of surgical data collection and annotation.
Surgical scenes are dynamic, anatomically intricate and safety-critical, requiring domain expertise to precisely characterize fine-grained details and procedural context. 
In particular, reasoning annotations demand multi-level clinical interpretation and careful validation, making large-scale consistent annotation substantially more challenging than standard perception labeling. 

In future iterations beyond Surg$\Sigma$-DB v0.1, we will continue expanding cross-task coverage and enriching structured reasoning annotation under a unified label space and annotation framework, moving toward more holistic and fully aligned multimodal surgical foundation training.

\section{Conclusion}
In this work, we introduced Surg$\Sigma$, a unified spectrum of large-scale multimodal data and foundation models for surgical intelligence. 
At its core, Surg$\Sigma$-DB provides a systematically curated and scalable multimodal data foundation, consolidating heterogeneous surgical resources into a unified schema with consistent semantics and standardized formats across diverse procedures. 
Surg$\Sigma$-DB supports comprehensive supervision spanning understanding, reasoning, planning, and generation tasks, and is designed around three key principles: large-scale multimodal data, unified data representations, and structured reasoning annotations. 
Empirical evidence from four foundation models built upon Surg$\Sigma$-DB demonstrates that these data-centric design choices substantially enhance cross-task generalization and interpretable reasoning in complex surgical environments. 
We envision Surg$\Sigma$-DB as a scalable infrastructure for surgical foundation modeling, and will continue expanding its scale, diversity, and annotation completeness toward dense cross-task coverage within unified surgical scenes, advancing clinically reliable and generalizable multimodal surgical intelligence.

\subsection*{Acknowledgments}
We sincerely thank Dillan Imans for his invaluable assistance with dataset annotation during his visit as a researcher at CUHK. We also thank Erli Zhang from NUS for his invaluable assistance with dataset summarizing.
{
\small
\bibliographystyle{plain}
\bibliography{reference}
}

\end{document}